\documentclass{article}

     \PassOptionsToPackage{numbers, compress}{natbib}
    \usepackage[preprint]{neurips_2024}

\usepackage[utf8]{inputenc} % allow utf-8 input
\usepackage[T1]{fontenc}    % use 8-bit T1 fonts
\usepackage{hyperref}       % hyperlinks
\usepackage{url}            % simple URL typesetting
\usepackage{booktabs}       % professional-quality tables
\usepackage{amsfonts}       % blackboard math symbols
\usepackage{nicefrac}       % compact symbols for 1/2, etc.
\usepackage{microtype}      % microtypography
\usepackage{multirow}
\usepackage{subcaption}
\usepackage{algorithm}
\usepackage{algorithmic}
\usepackage{amsmath}

\usepackage{graphicx}
\usepackage{booktabs}
\usepackage{bbding} % 导入宏包

\renewcommand{\mathbf}{\boldsymbol}
\def\x{\mathbf{x}}

\def\cc{\mathbf{c}}
\def\p{\mathbf{p}}
\def\C{\mathbf{C}}

\def\vv{\mathbf{v}}

\newcommand{\ie}{\emph{i.e.}}
\newcommand{\eg}{\emph{e.g.}}

\DeclareMathOperator*{\argmax}{arg\,max}
\usepackage[dvipsnames]{xcolor}
\usepackage{color, colortbl}

\definecolor{LightCyan}{rgb}{0.88,1,1}
\definecolor{HighLight}{rgb}{0.96,0.92,0.96}

\title{AdaNeg: Adaptive Negative Proxy Guided \\ OOD Detection with Vision-Language Models}
\author{%
  Yabin Zhang \\
  The Hong Kong Polytechnic University \\
  \texttt{csybzhang@comp.polyu.edu.hk} \\
  \And Lei Zhang\thanks{Corresponding Author.} \\
  The Hong Kong Polytechnic University \\
  \texttt{cslzhang@comp.polyu.edu.hk} \\
}

\begin{document}

\maketitle

\begin{abstract}

% The ability to detect out-of-distribution (OOD) samples is crucial for the reliability of AI models. Recent studies have demonstrated that pre-trained vision-language models possess significant capabilities for OOD detection, particularly when guided by negative texts. However, applying the same negative texts to all OOD datasets inevitably leads to semantic misalignment, where the negative texts do not align well with the label space of OOD images.
% To address this, we propose adaptive negative proxies by exploring potential OOD images during testing. These proxies are constructed with real OOD images, thus they align better with OOD label spaces and provide more effective guidance over negative labels.

%The detection of out-of-distribution (OOD) samples is crucial for the reliability of artificial intelligence models. 
Recent research has shown that pre-trained vision-language models are effective at identifying out-of-distribution (OOD) samples by using negative labels as guidance.
However, employing consistent negative labels across different OOD datasets often results in semantic misalignments, as these text labels may not accurately reflect the actual space of OOD images. To overcome this issue, we introduce \textit{adaptive negative proxies}, which are dynamically generated during testing by exploring actual OOD images, to align more closely with the underlying OOD label space and enhance the efficacy of negative proxy guidance.
Specifically, our approach utilizes a feature memory bank to selectively cache discriminative features from test images, representing the targeted OOD distribution. This facilitates the creation of proxies that can better align with specific OOD datasets.
While task-adaptive proxies average features to reflect the unique characteristics of each dataset, the sample-adaptive proxies weight features based on their similarity to individual test samples, exploring detailed sample-level nuances.
The final score for identifying OOD samples integrates static negative labels with our proposed adaptive proxies, effectively combining textual and visual knowledge for enhanced performance.
Our method is training-free and annotation-free, and it maintains fast testing speed.
Extensive experiments across various benchmarks demonstrate the effectiveness of our approach, abbreviated as AdaNeg. Notably, on the large-scale ImageNet benchmark, our AdaNeg significantly outperforms existing methods, with a 2.45\% increase in AUROC and a 6.48\% reduction in FPR95.
Codes are available at \url{https://github.com/YBZh/OpenOOD-VLM}.

\end{abstract}

\section{Introduction}

In real applications, artificial intelligence (AI) systems often encounter test samples of unknown classes, termed out-of-distribution (OOD) data. These OOD data often result in overly confident errors \cite{scheirer2012toward,nguyen2015deep}, posing security threats. Therefore, accurately identifying OOD data is essential for ensuring the reliability and security of AI systems in open-world environments.

%% 传统的 OOD detection 主要是基于vision-only model 来做，最近vision-language models have achieved 非凡的generalization performance by utilizing multi-modal knowledge. 具体来说，the text knowledge is introduced in OOD detection in XX. Recently, NegLabel 挖掘negative labels by exploring text labels that 远离ID labels. It achieved the state of -the art OOD detection performance by comparing the test image with provided ID labels and mined negative labels. 也就是说，NegLabel 把 mined negative texts 作为 OOD data 的代理. 但是这样的代理可能不是最优的，因为mined negative text 和 真实的 OOD images 不仅存在模态上的差异，也存在semantic shift, as shown in Figure XXX.  This shift between mined negative information and 真正的OOD images lead to 次优解。

\begin{figure}[tbp]
    \centering
    \begin{subfigure}[b]{0.35\linewidth}
        \centering
        \includegraphics[width=\linewidth]{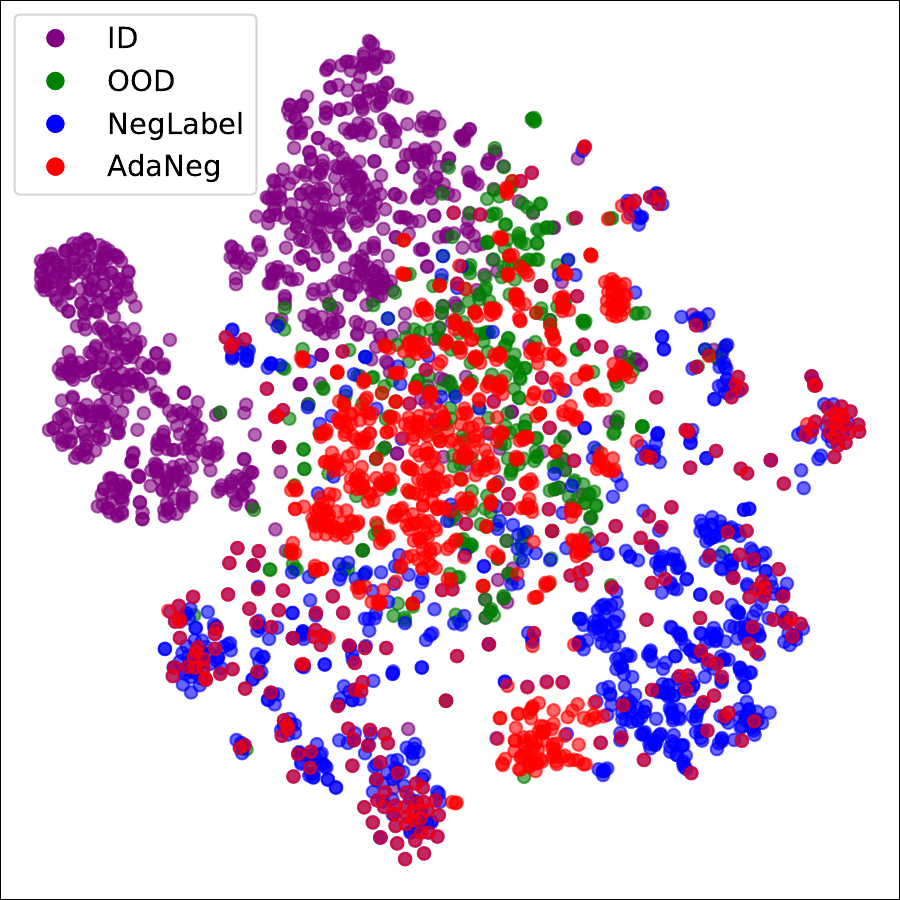}
        \caption{t-SNE Visualization}
        \label{fig:tsne_visualization_top1k}
    \end{subfigure}
        \hspace{0.02\linewidth}
    \begin{subfigure}[b]{0.35\linewidth}
        \centering
        \includegraphics[width=\linewidth]{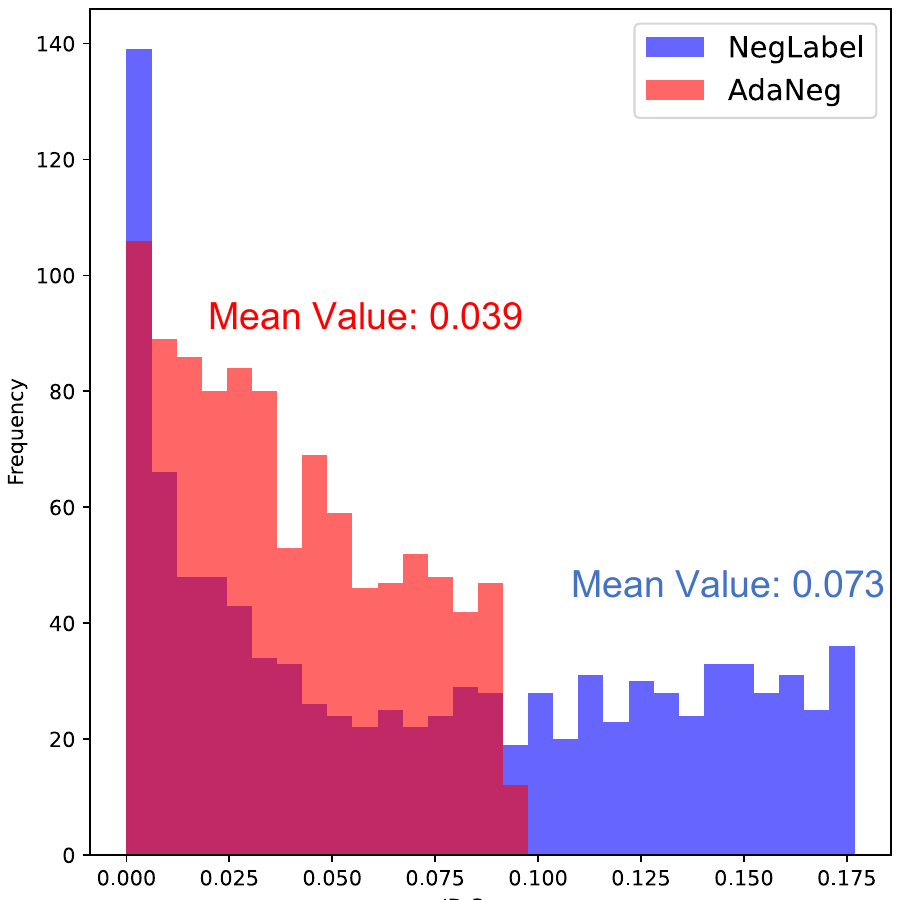}
        \caption{Histogram of ISOR}
        \label{fig:histogram_id_1k}
    \end{subfigure}
    \caption{Qualitative and quantitative analyses of semantic misalignment between OOD labels and negative proxies using ImageNet (ID) and SUN (OOD) datasets. (a) Visualization of ID labels, OOD labels, negative labels from NegLabel, and adaptive negative proxies (AdaNeg).
   (b) Quantitative analysis based on ID-Similarity to OOD Ratio (ISOR in short, see Appendix \ref{subsec:id_aligning_scores}). Lower ISOR indicates a higher similarity to OOD labels and reduced similarity to ID labels. AdaNeg consistently achieves lower ISOR, demonstrating enhanced alignment with OOD characteristics. Visualizations include the top 1,000 discriminative proxies from both NegLabel and AdaNeg.}
    \label{fig:histogram_tsne_misalignment}
    	\vspace{-0.2cm}
\end{figure}

Traditional OOD detection methods in image domain primarily rely on vision-only models \cite{hendrycks2016baseline,lee2018simple,liang2017enhancing}. Recent advancements in vision-language models (VLMs) have demonstrated remarkable OOD detection performance by leveraging multi-modal knowledge \cite{fort2021exploring,esmaeilpour2022zero,ming2022delving}. Recently, NegLabel \cite{jiang2024negative} explores negative labels by identifying text labels that are semantically distant from the in-distribution (ID) labels.
This method achieves state-of-the-art performance by detecting test images closer to negative labels as OOD. In other words, NegLabel regards these negative labels as proxies of OOD data. 
However, employing consistent negative labels across different OOD datasets often leads to semantic misalignment, where these text labels may not accurately reflect the actual label space of OOD images, as shown in Fig. \ref{fig:histogram_tsne_misalignment}. This misalignment between the proxies and targeted OOD distribution leads to sub-optimal performance.

To promote the alignment between the negative proxies and target OOD distribution, we introduce the \textbf{Ada}ptive \textbf{Neg}ative proxies (\textbf{AdaNeg}), which are dynamically generated during testing by exploring actual OOD images. Specifically, we start by initializing an empty category-split memory bank for each OOD dataset and selectively cache features of discriminative OOD images during testing. The OOD discrimination is assessed using mined negative labels, as detailed in \cite{jiang2024negative}.
With this feature memory, we develop task-adaptive proxies by simply averaging cached features within each category. These proxies, derived from actual OOD images, reflect the distinct characteristics of the target OOD dataset and align more closely with the underlying OOD label space.

The task-adaptive proxies mentioned previously provide unique proxies for different OOD datasets while maintaining consistency across various test samples within the same dataset.
To delve into the fine-grained nuances at the sample level, we introduce the sample-adaptive proxies by weighting cached features based on their similarity to a particular test sample. This is achieved with an attention mechanism, where the feature memory serves as both keys and values, and the test feature acts as the query. The final score for detecting OOD samples integrates static negative labels with our adaptive proxies, effectively combining textual and visual knowledge for enhanced performance.

We conduct extensive experiments on standard benchmarks to validate the effectiveness of AdaNeg, where our proposed adaptive proxies outperform the negative-label-based one, enhancing performance with complementary multi-modal knowledge. 
Particularly, on the large-scale ImageNet dataset, our AdaNeg method outperforms existing methods by 2.45\% AUROC and 6.48\% FPR95. Notably, our method is training-free and annotation-free, and it maintains fast testing speed, as analyzed in Tab. \ref{Tab:computation_efficiency_adaneg}.
The ability to dynamically adjust to new OOD datasets without affecting testing speed or labor-intensive annotation/training makes our approach particularly valuable for real-world applications where adaptability and efficiency are crucial.
We summarize our contribution as follows:
\begin{itemize}
    \item We first identify the label space misalignment between existing negative-label-based proxies and the target OOD distributions. In response, we introduce adaptive negative proxies that are dynamically generated during testing by exploring actual OOD images, resulting in a more effective alignment with the OOD label space.
    \item Our adaptive negative proxies are constructed with a feature memory bank that selectively caches discriminative image features during testing. We instantiate this concept by developing task-adaptive proxies to reflect the unique characteristics of each OOD dataset and sample-adaptive proxies to capture detailed sample-level nuances. The final OOD detection score combines these insights with complementary textual and visual knowledge.
    \item We conduct thorough analyses of the proposed components and perform extensive experiments on standard benchmarks. Our method is training-free and annotation-free, and it maintains fast testing speed and achieves state-of-the-art performance. Notably, our method significantly outperforms existing methods, with a 2.54\% increase in AUROC and a 6.48\% reduction in FPR95 on the large-scale ImageNet dataset.
\end{itemize}

\section{Related Work}

\textbf{OOD Detection} focuses on identifying OOD test samples with semantic shifts, thus distinguishing it from generalization studies which typically focus on covariate shifts \cite{chen2022relation,chen2023coda,chen2023activate,zhang2020unsupervised}.
A variety of OOD detection techniques have been developed, which can be roughly categorized into score-based \cite{hendrycks2016baseline, lee2018simple, liang2017enhancing, liu2020energy, wang2021energy, huang2021mos, wang2022vim, sun2021react}, distance-based \cite{tack2020csi, tao2023non, sun2022out, du2022siren, ming2022exploit, sehwag2021ssd}, and generative-based \cite{ryu2018out, kong2021opengan} methods.
Among them, score-based methods are particularly notable by employing a variety of scoring functions to differentiate between ID and OOD samples. These functions include confidence-based \cite{hendrycks2016baseline, liang2017enhancing, sun2021react, wang2022vim}, discriminator-based \cite{kong2021opengan}, energy-based \cite{liu2020energy, wang2021energy}, and gradient-based \cite{huang2021importance} scores. In contrast, distance-based methods determine OOD samples by evaluating the distance in the feature space between test data and the closest ID samples \cite{tack2020csi} or ID prototypes \cite{tao2023non}, using metrics such as KNN \cite{sun2022out, du2022siren, ming2022exploit} or Mahalanobis distance \cite{lee2018simple, sehwag2021ssd}.

Despite their achievements, traditional OOD detection methods generally rely on manually annotated ID images and often overlook the integration of textual information. To leverage the textual knowledge, recent advancements have focused on employing VLMs \cite{ming2022delving, ming2024does, jiang2024negative, zhang2024dual,zhang2024lapt,esmaeilpour2022zero, miyai2024locoop, wang2023clipn, nie2023out}.
Specifically, ZOC \cite{esmaeilpour2022zero} applies VLMs to discern OOD instances by training a captioner that generates potential OOD labels. Nevertheless, this captioner often fails to produce effective OOD labels, particularly for ID datasets containing many classes.
LoCoOp \cite{miyai2024locoop} adopts a novel approach by learning ID prompts from few-shot ID samples, and further enhances the robustness of these prompts by incorporating OOD features mined from the backgrounds of images.
CLIPN \cite{wang2023clipn}, LSN \cite{nie2023out} and LAPT \cite{zhang2024lapt} explore learning text prompts for expressing negative concepts. In specific, CLIPN initializes text prompts with the word `no' combined with ID labels and refines them with large-scale multi-modal data; LSN starts with manually collected ID samples to learn negative prompts, offering a different approach to leveraging textual information in OOD detection; LAPT conducts automated prompt tuning with automatically collected training samples, boosting OOD detection without any manual effort.
MCM \cite{ming2022delving} utilizes ID class names to facilitate effective zero-shot OOD detection. It is further refined by NegLabel \cite{jiang2024negative}, which incorporates additional negative class names mined from available data sources as negative proxies. However, as illustrated in Fig. \ref{fig:histogram_tsne_misalignment}, there is a mismatch between the negative-label-based proxies and the target OOD distribution, underscoring the limitations of this strategy. This observation has inspired us to construct adaptive proxies by exploring potential OOD test images during testing. This leads to an efficient method that aligns better with the target OOD distribution, resulting in enhanced OOD detection performance.

Furthermore, we clarify the relationship between our method and existing approaches on OOD exposure \cite{fort2021exploring, hendrycks2018deep, zhang2021fine}. Most OOD exposure methods introduce manually collected negative images during training, where manual labor is necessary to ensure that the labels of negative images are different from ID ones. Moreover, involving negative images in training typically introduces additional computational overhead, impeding its practical deployment.
Unlike these methods, NegLabel \cite{jiang2024negative} is exposed to negative labels during the test phase in a training-free manner. However, given a fixed ID dataset, the exposed negative texts remain consistent for different OOD datasets, inevitably resulting in label misalignment, as shown in Fig. \ref{fig:histogram_tsne_misalignment}. 
To address this, we expose the VLMs to adaptive negative proxies, which explore actual OOD samples during testing and align more effectively with OOD distribution. Our method does not require manual annotations and works in a training-free manner, making it an appealing solution for real applications.

%%%%%%%%%%%%%%%%% 新出的几篇论文也加进来。 CVPR24 以及ICCV的。s
\textbf{Test-time Adaptation.}
We adopt an online update of the negative proxies during testing,  resembling test-time adaptation (TTA) methods \cite{liang2023comprehensive,wang2020tent,shu2022test}. Existing TTA methods primarily address covariate shifts between training and testing domains. In contrast, our approach mitigates the label shift between negative proxies and the target OOD distribution by exploring online test samples.
Recently, TTA strategies have been considered in the field of OOD detection. However, these methods typically require test-time optimization \cite{gao2023atta,yang2023auto,fan2022simple}, slowing down the testing process. In contrast, our method is optimization-free and introduces only a lightweight memory interaction operation, enabling rapid and accurate testing, as analyzed in Tab. \ref{Tab:computation_efficiency_adaneg}.

\textbf{Memory Networks.} 
The use of memory networks for storing and retrieving past knowledge \cite{weston2014memory, sukhbaatar2015end} has been extensively applied across various fields, including classification \cite{karunaratne2021robust,santoro2016meta,zhang2024dual}, segmentation \cite{oh2019video,xie2021few}, detection \cite{deng2019object,chen2020memory}, and NLP \cite{packer2023memgpt}. To our best knowledge, our work is the first to apply memory networks to the field of OOD detection. By caching and retrieving test images with a feature memory, we propose adaptive proxies to more effectively align with the OOD distribution in a training-free manner. This innovative approach significantly enhances the OOD detection performance.

% The strategy of storage and retrieval of past knowledge with memory networks \cite{weston2014memory,sukhbaatar2015end} has been widely used in various fields, such as classification \cite{xx}, segmentation \cite{xx}, detection \cite{xx}, and NLP \cite{xx}. To our knowledge, we are the first to explore memory networks in OOD detection. By caching and retrievaling test images with a feature memory, we introduce adaptive proxies that align more effectively with OOD distribution in a training-free manner, resulting in boosted OOD detection results.

\begin{figure}
	\centering
	\includegraphics[width=0.86\linewidth]{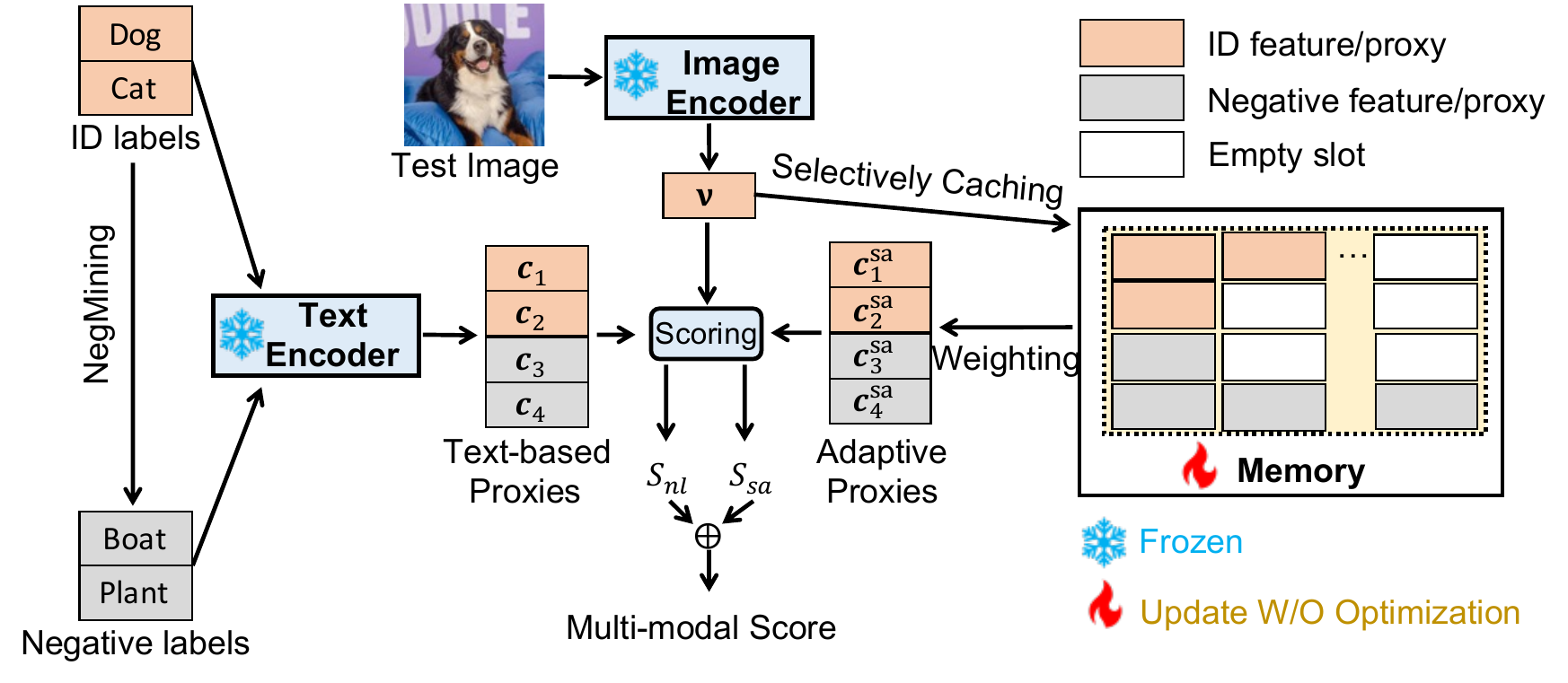}
		\vspace{-0.1cm}
	\caption{The overall framework of AdaNeg, where we selectively cache test images and generate adaptive proxies with an external feature memory bank. The final score combines textual and visual knowledge from static negative labels and our adaptive proxies, integrating multi-modal information. } \label{Fig:framework_adaneg}
	\vspace{-0.2cm}
\end{figure}

\vspace{-0.1cm}
\section{Methodology} \label{sec:method_adaneg}
	\vspace{-0.1cm}
\subsection{Preliminaries}
	\vspace{-0.1cm}

\textbf{OOD Detection Setup.}
Consider $\mathcal{X}$ as the image domain and $\mathcal{Y} = \{ y_1, \dots, y_C \}$ as the space of ID class labels, where $\mathcal{Y}$ comprises text elements such as $\mathcal{Y} = \{ cat, dog, \dots, bird \}$, and $C$ represents the total number of classes. Let $\x^{in}$ and $\x^{ood}$ be random variables representing ID and OOD samples from $\mathcal{X}$, respectively. We define $\mathcal{P}{\x^{in}}$ and $\mathcal{P}{\x^{ood}}$ as the marginal distributions for ID and OOD, respectively.
In conventional classification scenarios, it is assumed that the test image $\x$ originates from the ID and is associated with a specific ID label, specifically, $\x \in \mathcal{P}{\x^{in}}$ and $y \in \mathcal{Y}$, with $y$ being the label of $\x$.
However, in real applications, AI systems often face data from unknown classes, denoted by $\x \in \mathcal{P}{\x^{ood}}$ and $y \notin \mathcal{Y}$. Such occurrences can make AI models incorrectly categorize these instances into familiar ID categories with substantial certainty \cite{scheirer2012toward,nguyen2015deep}, resulting in security concerns.
To address these challenges, OOD detection is proposed to accurately categorize ID samples into their respective classes and reject OOD samples as non-ID. Recognition within the ID categories is performed using a $C$-way classifier, following standard classification approaches \cite{krizhevsky2012imagenet,he2016deep}. Concurrently, OOD detection typically employs a scoring mechanism $S$ \cite{lee2018simple,liang2017enhancing,liu2020energy} to differentiate between ID and OOD inputs:
\begin{equation}
    G_{\gamma}(\x) = \textrm{ID, } \textrm{if } S(\x) \geq \gamma; \quad \textrm{otherwise, } G_{\gamma}(\x) = \textrm{OOD},
\end{equation}
where $G_{\gamma}$ represents the OOD detector set at a threshold $\gamma \in \mathcal{R}$. The test sample $\x$ is identified as an ID sample if and only if $S(\x) \geq \gamma$.

\textbf{CLIP and NegLabel.}
For an ID test image $\x$ within the label space $\mathcal{Y}$, we derive the image feature vector $\vv = f_{img}(\x) \in \mathcal{R}^D$ and the text feature matrix $\C_{id} = f_{txt}(\rho(\mathcal{Y})) \in \mathcal{R}^{C\times D}$ using pre-trained CLIP encoders, where $D$ represents the feature dimension. The functions $f_{img}(\cdot)$ and $f_{txt}(\cdot)$ are the encoders for images and text, respectively.
The function $\rho(\cdot)$ is the text prompt mechanism, typically defined as `a photo of a <label>.', where label is the actual class name, for example, `cat' or `dog'.
Both $\vv$ and $\C_{id}$ undergo $L_2$ normalization across the dimension $D$. Then, zero-shot classification probabilities are computed utilizing $\C$ as the classifier:
\begin{equation}
    \p^{id} = \mathrm{Softmax}(\vv\C_{id}^T/\tau) \in \mathcal{R}^{C},
\end{equation}
where $\tau > 0$ is the scaling temperature. 

The vanilla CLIP is proposed for zero-shot ID recognition and has recently been extended to OOD detection. Specifically, the NegLabel approach \cite{jiang2024negative} introduces negative class names $\mathcal{Y}^- =\{ y_{C+1}, \dots, y_{C+M} \}$ sourced from broad text corpora, where $M$ is the length of negative classes and $\mathcal{Y}^- \cap \mathcal{Y} = \emptyset$. 
Then, we can obtain the full text feature matrix $\C = f_{txt}(\rho(\mathcal{Y}\cup\mathcal{Y}^-)) \in \mathcal{R}^{(C+M)\times D}$ with the pre-trained CLIP text encoder, leading to the classification probability across $C+M$ classes:
\begin{equation}
    \p = \mathrm{Softmax}(\vv\C^T/\tau) \in \mathcal{R}^{C+M}.
\end{equation}
Assuming that ID samples exhibit greater similarity to ID labels and lesser similarity to negative labels compared to OOD samples, NegLabel introduces the following score for OOD detection:
\begin{equation} \label{equ:neglabel_score}
    S_{nl}(\vv) =  \sum_{i=1}^C \p_i,
\end{equation}
where $\p_i$ is the $i$-th entry of $\p$, indicating the classification probability of the $i$-th class.
Intuitively, the NegLabel method uses negative labels as proxies of the OOD distribution, detecting OOD images based on the similarity to these negative labels.

% \begin{equation} \label{equ:neglabel_score}
%     S_{NL}(\vv) =  \frac{\sum_{i=1}^C {e^{\cos(\vv, \cc_i)/\tau}}}{\sum_{i=1}^C {e^{\cos(\vv, \cc_i)/\tau}} + \sum_{j=C+1}^{C+M} {e^{\cos(\vv, \cc_j)/\tau}}},
% \end{equation}
% where $\mathrm{cos}(\cdot,\cdot)$ measures the cosine similarity, $\cc_i$ is the $i$-th entry of $\C$.
%NegLabel achieves impressive results across multiple OOD detection benchmarks by leveraging knowledge from ID and the explored negative labels.

\subsection{AdaNeg}

Although NegLabel has successfully employed negative labels as the OOD proxies, there is a semantic misalignment between such OOD proxies and actual OOD labels, as illustrated in Fig. \ref{fig:histogram_tsne_misalignment}.
We aim to obtain OOD proxies that align better to the targeted OOD distribution. However, acquiring such OOD proxies is challenging, as the OOD information is unknown prior to actual testing.

From another perspective, we can access real OOD information during testing, motivating us to refine OOD proxies in the testing stage. We can identify discriminative negative images during testing and then adjust the OOD proxies toward detected images. This is achieved by selectively caching features of test images into a task-aware memory bank, followed by memory reading operations to produce adaptive proxies. We detail our implementation as follows.

\textbf{Task-aware Memory.}
We construct a task-aware memory as a category-split tensor $\mathbf{M} \in \mathbb{R}^{(C+M)\times L \times D}$, where $L$ is the memory length for each category. $\mathbf{M}$ is initialized with zero values and gradually filled with features of selected images during testing.  
Specifically, for a test image with feature $\vv$, we first calculate its score $S_{nl}(\vv)$ with Eq. (\ref{equ:neglabel_score}). If $S_{nl}(\vv) < \gamma$, we detect this test point as a negative image, and otherwise, it is identified as a positive sample. For negative and positive images, we respectively identify their closest labels as:
\begin{align} 
\mathrm{Negative}: y =& \argmax_i \p^{ood}_i +C,  \label{Equ:pseudo_labeling_ood} \\
\mathrm{Positive}: y =& \argmax_i \p^{id}_i, \label{Equ:pseudo_labeling_id}
\end{align}
where $\p^{ood} = \p[C: C+M]$ and $\p^{id} = \p[0:C]$ are the classification probabilities corresponding to negative and ID class names, respectively.
 Then, we cache this feature $\vv$ into the task-aware memory:
\begin{equation} \label{Equ:feat_cache_ood}
\mathbf{M}_{y, l} = \vv,
\end{equation}
where $l$ indicates an empty slot of $\mathbf{M}_y \in \mathbb{R}^{L \times D}$. Once $\mathbf{M}_y$ is filled, we drop the image feature with the highest prediction entropy, as detailed in Appendix \ref{subsec:entropy_based_caching}.  In one word, we keep confident image features with low prediction entropy in the memory.

%One may opt to replace Eqs. \ref{Equ:pseudo_labeling_ood} and \ref{Equ:pseudo_labeling_id} with a simpler and unified one: $y = \argmax_i \p_i$, which will reduce the distinguishing between ID and OOD samples \textcolor{red}{as analyzed in Sec. XXX.}  

The aforementioned strategy attempts to cache all test images, including those with high confusion between ID and OOD.
However, we found that selectively retaining only those image features that exhibit strong ID/OOD distinguishing capabilities can effectively reduce this confusion. Specifically, we modify the selection criterion for memorization as follows:
\begin{align}  \label{Equ:criterion_memorized_tau_g}
\mathrm{Negative}: S_{nl}(\vv) < \gamma &\to S_{nl}(\vv) < \gamma - g\gamma , \notag  \\
\mathrm{Positive}: S_{nl}(\vv) \geq  \gamma &\to S_{nl}(\vv) \geq  \gamma + g(1-\gamma), 
\end{align}
where $g \in [0,1]$ is a hyperparameter that introduces a gap in the score space. Consequently, image features falling within the gap $\gamma - g\gamma \leq S_{nl}(\vv) < \gamma + g(1-\gamma)$ are considered to have low distinguishing confidence and are not cached.

\textbf{Task-adaptive Proxies.}
Given the updated $\mathbf{M}$, we can easily get the task-adaptive proxy by averaging along the length dimension $L$:
\begin{equation} \label{Eq:task_adaptive_classifier}
    \C^{ta} =   \mathrm{L_2}\left( \frac{1}{L+1} \sum_{l=1}^{L+1} \widehat{\mathbf{M}}_{:,l,:} \right) \in \mathbb{R}^{(C+M)\times D},
\end{equation}
where $\mathrm{L_2}(\cdot)$ indicates the $L_2$ normalization along feature dimension $D$, and $\widehat{\mathbf{M}} = [\mathbf{M}, \mathbf{C} ] \in \mathbb{R}^{(C+M)\times (L+1)\times D}$ is the extended memory with vanilla text proxies $\C$. Such a memory extension is necessary since $\mathbf{M}$ is initially empty and uninformative. Initially, this extension initializes the adaptive proxies $\C^{ta}$ with the basic text proxies $\C$.
However, there are two key distinctions between the adaptive $\C^{ta}$ and the vanilla $\C$. First, unlike the NegLabel approach, which employs a fixed proxy $\C$ across various OOD datasets, our $\C^{ta}$ dynamically adjusts to the targeted OOD domain as the memory accumulates data, thereby providing dataset-specific adaptive proxies. 
 Second, the $\C^{ta}$ primarily incorporates image features, offering modal knowledge that complements the text-based proxies $\C$. The benefits of this approach are further analyzed in Tab. \ref{Tab:ablation_score_fun}.

% Second, the proposed $\C^{ta}$ is mainly constructed with image features, providing modally complementary knowledge to the text-feature-based $\C$, as analyzed in Tab. \ref{Tab:ablation_score_fun}. 

The score function for OOD detection with the task-adaptive proxy $\C^{ta}$ is derived as:
\begin{equation} \label{equ:adaneg_score_taskada}
    S_{ta}(\vv) =  \frac{\sum_{i=1}^C {e^{\cos(\vv, \cc^{ta}_i)/\tau}}}{\sum_{i=1}^C {e^{\cos(\vv, \cc^{ta}_i)/\tau}} + \sum_{j=C+1}^{C+M} {e^{\cos(\vv, \cc^{ta}_j)/\tau}}},
\end{equation}
where $\mathrm{cos}(\cdot,\cdot)$ measures the cosine similarity, and $\cc^{ta}_i$ is the $i$-th entry of $\C^{ta}$.

\textbf{Sample-adaptive Proxies.}
As evidenced in Table \ref{Tab:ablation_score_fun}, our task-adaptive proxies significantly outperform the fixed text proxies by effectively adapting to the characteristics of target OOD dataset. Building on this success, we further refine our approach by leveraging finer-grained, sample-level nuances to introduce even more effective sample-adaptive proxies.
 Specifically, given the extended memory $\widehat{\mathbf{M}}$ and the test image feature $\vv$, we introduce the sample-adaptive proxy $\C^{sa} \in \mathbb{R}^{(C+M)\times D}$ via the following cross-attention operation:
\begin{align} \label{Equ:sample_adaptive_classifier}
\cc^{sa}_i =  \mathrm{L_2}\left( \varphi\left(\vv(\widehat{\mathbf{M}}_i)^{\top} \right)\widehat{\mathbf{M}}_i \right) \in \mathbb{R}^{D},
\end{align}
where $\vv(\widehat{\mathbf{M}}_i)^{\top} \in \mathbb{R}^{1\times (L+1)}$ measures the cosine similarities between normalized features of $\vv$ and $\widehat{\mathbf{M}}_i$, and $\varphi(x) = \exp(-\beta (1-x))$ modulates the sharpness of $x$ with hyper-parameter $\beta$. $\cc^{sa}_i$ and $\widehat{\mathbf{M}}_i$ are the $i$-th entry of $\C^{sa}$ and $\widehat{\mathbf{M}}$, respectively.

\begin{algorithm} [tb]
\caption{Adaptive Negative Proxy Guided OOD Detection} \label{algorithm:adaneg}
\begin{algorithmic}[1] 
    \REQUIRE ID label space $\mathcal{Y}$ and test set $\mathcal{X}$ % Input conditions
    \STATE $\mathcal{Y}^- \leftarrow$ NegMine($\mathcal{Y}$) following \cite{jiang2024negative}
    \STATE Constructing an empty memory $\mathbf{M}$
    \FOR{$\x \in \mathcal{X}$} 
        \STATE Generating detection score with negative labels using Eq. \ref{equ:neglabel_score}
        \STATE Determine whether $\x$ should be cached using Eq. \ref{Equ:criterion_memorized_tau_g}
        \STATE Caching $\x$ with Eq. \ref{Equ:feat_cache_ood} if needed
        \STATE Generating adaptive proxies with memory bank using Eq. \ref{Eq:task_adaptive_classifier} or Eq. \ref{Equ:sample_adaptive_classifier}
        \STATE Generating adaptive scores with Eq. \ref{equ:adaneg_score_taskada} or Eq. \ref{equ:adaneg_score_sampleada} 
        \STATE Generating final score $S_{all}$ by merging multi-modal knowledge with Eq. \ref{equ:adaneg_score_all}
    \ENDFOR
    \STATE \textbf{Return} Collected final scores \{$S_{all}$\}
    % \RETURN Collected final scores $S_{all}$ 
\end{algorithmic}
\end{algorithm}

Both the task-adaptive and the sample-adaptive proxies are derived from the memorized image features stored in $\widehat{\mathbf{M}}$. The primary distinction between them lies in their respective weighting strategies. For $\C^{ta}$, each feature $\widehat{\mathbf{M}}_{:,l,:}$ in the memory is assigned a uniform weighting coefficient of $\frac{1}{L+1}$. Conversely, in constructing $\C^{sa}$, the weighting coefficient for each cached feature is dynamically determined based on its cosine similarity to the test image feature, denoted as $\vv(\widehat{\mathbf{M}}_i)^{\top}$. Consequently, while $\C^{ta}$ remains constant across different test samples, $\C^{sa}$ adapts to each individual test sample. This adaptability allows $\C^{sa}$ to tailor its response based on the specific characteristics of each test image, thereby enhancing discrimination between ID and OOD samples, particularly in diverse and variable testing scenarios.

%Essentially, both the task-adaptive and the sample-adaptive proxies are derived from the memorized image features in $\widehat{\mathbf{M}}$, with the main difference between them lying in their weighting strategies. Specifically, in the construction of $\C^{ta}$, each cached feature of $\widehat{\mathbf{M}}_i$ is given the same weighting coefficient, $\frac{1}{L+1}$. However, in the construction of $\C^{sa}$, the weighting coefficient for each cached feature is adaptively determined based on its cosine similarity with the test image feature, \ie, $\vv(\widehat{\mathbf{M}}_i)^{\top}$. Thus, given the same memory, which contains task-specific image features, $\C^{ta}$ is fixed for different test samples, whereas $\C^{sa}$ dynamically adapts for different test samples. This adaptability explores specific characteristics of each test sample and enables more precise discrimination between ID and OOD samples, particularly in diverse and variable testing environments.

The score function for OOD detection with the sample-adaptive proxies $\C^{sa}$ is derived as:
\begin{equation} \label{equ:adaneg_score_sampleada}
    S_{sa}(\vv) =  \frac{\sum_{i=1}^C {e^{\cos(\vv, \cc^{sa}_i)/\tau}}}{\sum_{i=1}^C {e^{\cos(\vv, \cc^{sa}_i)/\tau}} + \sum_{j=C+1}^{C+M} {e^{\cos(\vv, \cc^{sa}_j)/\tau}}}.
\end{equation}

\textbf{Multi-modal Score.}
As previously discussed, the score function $S_{nl}(\vv)$ relies primarily on text features, whereas the sample-adaptive score function $S_{sa}(\vv)$ utilizes cached image features. Given the complementary nature of text and image modalities, we derive the final score function by integrating knowledge from both modalities:
\begin{equation} \label{equ:adaneg_score_all}
    S_{all}(\vv) =  S_{nl}(\vv) + \lambda S_{sa}(\vv),
\end{equation}
where $\lambda > 0$ is the hyperparameter balancing the two modalities. The overall pipeline of our method is illustrated in Fig. \ref{Fig:framework_adaneg} and summarized in Algorithm \ref{algorithm:adaneg}.

\section{Experiments}
\subsection{Setup} \label{subsec:exp_setup_adaneg}
	\vspace{-0.1cm}
	
\textbf{Datasets.}
We conduct extensive experiments with the large-scale ImageNet-1k \cite{deng2009imagenet} as ID data. Following prior practice \cite{huang2021mos,ming2022delving,jiang2024negative}, four OOD datasets of iNaturalist \cite{van2018inaturalist}, SUN \cite{xiao2010sun}, Places \cite{zhou2017places}, and Textures \cite{cimpoi2014describing} are evaluated. 
We also validate our method on the OpenOOD benchmark \cite{zhang2023openood,yang2022openood}, where OOD datasets are grouped into near-OOD (\eg, SSB-hard \cite{vaze2021open}, NINCO \cite{bitterwolf2023ninco}) and far-OOD (\eg, iNaturalist \cite{van2018inaturalist}, Textures \cite{cimpoi2014describing}, OpenImage-O \cite{wang2022vim}) according to their similarity to ImageNet dataset.
Besides ImageNet, we also evaluate our method on smaller-sized CIFAR10/100 datasets \cite{krizhevsky2009learning} with the OpenOOD setup. Specifically, with the ID dataset of CIFAR10/100, we adopt near-OOD datasets of CIFAR100/10 and TIN \cite{le2015tiny}, and far-ood datasets of MNIST \cite{deng2012mnist}, SVHN \cite{netzer2011reading}, Texture \cite{cimpoi2014describing}, and Plances365 \cite{zhou2017places}.
These experiments with various ID and OOD datasets enable a comprehensive evaluation on various OOD settings. 

% Our validation primarily uses the large-scale ImageNet-1k dataset \cite{deng2009imagenet} as ID data, and more evaluations on smaller-scaled ID datasets are provided in the Supplementary Materials. Aligning with standards from prior research \cite{huang2021mos,ming2022delving,jiang2024negative}, we use four diverse datasets of iNaturalist \cite{van2018inaturalist}, SUN \cite{xiao2010sun}, Places \cite{zhou2017places}, and Textures \cite{cimpoi2014describing} as OOD test sets.
% Additionally, we assess our method on the OpenOOD benchmark \cite{zhang2023openood,yang2022openood}, which differentiates OOD datasets into near-OOD (e.g., SSB-hard \cite{vaze2021open}, NINCO \cite{bitterwolf2023ninco}) and far-OOD (e.g., iNaturalist \cite{van2018inaturalist}, Textures \cite{cimpoi2014describing}, OpenImage-O \cite{wang2022vim}) based on semantic similarity or difficulty. This grouping allows for a comprehensive evaluation of OOD detectors against various OOD sample types.

\textbf{Implementation Details.}
We adopt the visual encoder of VITB/16 pretrained by CLIP \cite{radford2021learning} and analyze more backbone architectures in Tab. \ref{tab:ood_various_arch_adaneg}. For hyper-parameters, we adopt the memory length $L$=10, threshold $\gamma$=0.5 with the gap $g$=0.5 in Eq. \ref{Equ:criterion_memorized_tau_g}, $\beta$=5.5 in Eq. \ref{Equ:sample_adaptive_classifier}, and $\lambda$=0.1 in Eq. \ref{equ:adaneg_score_all} in all experiments. These hyper-parameters are carefully analyzed in Sec. \ref{subsec:adaneg_analyses}.
Following NegLabel, we adopt the text prompt of `The nice <label>.', set temperature $\tau$=0.01, and define the number $M$ of negative labels as $10,000$ for the ImageNet dataset. For the CIFAR datasets, we set the number $M$ as $70,000$ since we find that a larger $M$ leads to better results for CIFAR.
Following common practice \cite{huang2021mos,ming2022delving,jiang2024negative}, we employ the evaluation metrics of FPR95, AUROC, and ID ACC, which are detailed in Appendix \ref{subsec:criterion_definition}.
All experiments are conducted with a single Tesla V100 GPU.

\subsection{Main Results}
\vspace{-0.1cm}
\textbf{ImageNet Results with Four OOD Datasets.}
As illustrated in Tab. \ref{tab:four_ood_datasets_adaneg}, our AdaNeg significantly outperforms existing training-free methods and even surpasses approaches requiring additional training. Specifically, we report traditional methods \cite{hendrycks2016baseline,liang2017enhancing,liu2020energy,huang2021importance,wang2022vim,sun2022out,du2022unknown,tao2023non} by fine-tuning CLIP-encoders with labeled training samples following \cite{jiang2024negative}, and report results of \cite{nie2023out,miyai2024locoop,jiang2024negative,li2024learning,bai2023id} from their original paper.
Compared to the closest competitor \cite{jiang2024negative},  our method achieves consistent and notable improvements, validating the advantage of the proposed adaptive proxies over the negative-label-based ones.

\textbf{ImageNet Results with OpenOOD Setup.}
Our method is extensively evaluated against a range of OOD datasets in Tab. \ref{tab:openood_imagenet_adaneg} on the OpenOOD benchmark. Competing methods that require training are referred from OpenOOD. These methods utilize the full ImageNet training set and hold an unfair advantage over zero-shot, training-free methods like ours. Our AdaNeg consistently outperforms its closest competitor \cite{jiang2024negative} in both near-OOD and far-OOD scenarios. Additionally, AdaNeg not only enhances OOD detection capabilities but also improves ID classification, presenting higher robustness under diverse conditions.

Results on CIFAR10/100 datasets are provided in Appendix \ref{subsec:cifar_results}, where our advantages still hold.
%\textbf{CIFAR10/100 Results with OpenOOD Setup.}
%As illustrated in Tab. \ref{tab:openood_cifar_adaneg}, the advantage of our AdaNeg also holds on the CIFAR10/100 dataset. Notably, our method achieves new state-of-the-art results in the far-OOD setting under a zero-shot training-free manner, even outperforming competitors training on the full labeled training set.

% \clearpage 

\begin{table}[tb] \scriptsize
\centering
\caption{OOD detection results with ImageNet-1k, where a VITB/16 CLIP encoder is adopted.} \label{tab:four_ood_datasets_adaneg}
% \vspace{-0.2cm}
\begin{tabular}{lcccccccc|cc}
\toprule
\multicolumn{11}{c}{OOD datasets}  \\
\multicolumn{1}{c}{\multirow{2}{*}{Methods}} & \multicolumn{2}{c}{INaturalist} & \multicolumn{2}{c}{Sun} & \multicolumn{2}{c}{Places} & \multicolumn{2}{c}{Textures} & \multicolumn{2}{c}{Average} \\ \cline{2-3} \cline{4-5} \cline{6-7} \cline{8-9} \cline{10-11}
 & \tiny AUROC$\uparrow$ & \tiny FPR95$\downarrow$& \tiny AUROC$\uparrow$ & \tiny FPR95$\downarrow$& \tiny AUROC$\uparrow$ & \tiny FPR95$\downarrow$& \tiny AUROC$\uparrow$ & \tiny FPR95$\downarrow$ & \tiny AUROC$\uparrow$ & \tiny FPR95$\downarrow$        \\
 \midrule
 \multicolumn{11}{c}{\textbf{Methods requiring training (or fine-tuning)}} \\
MSP \cite{hendrycks2016baseline}    &    87.44 & 58.36 & 79.73 & 73.72 & 79.67 & 74.41 & 79.69 & 71.93 & 81.63 & 69.61   \\
ODIN \cite{liang2017enhancing}   &    94.65 & 30.22 & 87.17 & 54.04 & 85.54 & 55.06 & 87.85 & 51.67 & 88.80 & 47.75   \\
Energy \cite{liu2020energy}  &    95.33 & 26.12 & 92.66 & 35.97 & 91.41 & 39.87 & 86.76 & 57.61 & 91.54 & 39.89 \\
GradNorm \cite{huang2021importance} &   72.56 & 81.50 & 72.86 & 82.00 & 73.70 & 80.41 & 70.26 & 79.36 & 72.35 & 80.82 \\
ViM \cite{wang2022vim}   &    93.16 & 32.19 & 87.19 & 54.01 & 83.75 & 60.67 & 87.18 & 53.94 & 87.82 & 50.20 \\
KNN \cite{sun2022out}   &    94.52 & 29.17 & 92.67 & 35.62 & 91.02 & 39.61 & 85.67 & 64.35 & 90.97 & 42.19 \\
VOS \cite{du2022unknown}   &    94.62 & 28.99 & 92.57 & 36.88 & 91.23 & 38.39 & 86.33 & 61.02 & 91.19 & 41.32 \\
NPOS \cite{tao2023non}  &    96.19 & 16.58 & 90.44 & 43.77 & 89.44 & 45.27 & 88.80 & 46.12 & 91.22 & 37.93\\
ZOC \cite{esmaeilpour2022zero}               & 86.09 & 87.30 & 81.20 & 81.51 & 83.39 & 73.06 & 76.46 & 98.90 & 81.79 & 85.19 \\
LSN \cite{nie2023out} & 95.83 & 21.56 & 94.35 & 26.32 & 91.25 & 34.48 & 90.42 & 38.54 & 92.96 & 30.22 \\
CLIPN \cite{wang2023clipn}            & 95.27 & 23.94 & 93.93 & 26.17 & 92.28 & 33.45 & 90.93 & 40.83 & 93.10 & 31.10 \\
LoCoOp \cite{miyai2024locoop} & 96.86 & 16.05 & 95.07 & 23.44 & 91.98 & 32.87 & 90.19 & 42.28 & 93.52 & 28.66 \\
% Bai \emph{et al.} \cite{bai2023id} & 99.60 & 1.62 & 93.70 & 29.53 & 96.05 & 20.81 & 97.30 & 13.55 & 96.66 & 16.38 \\
LAPT \cite{zhang2024lapt}  & 99.63 & 1.16 & 96.01 & 19.12 & 92.01 & 33.01 & 91.06 & 40.32 & 94.68 & 23.40 \\
NegPrompt \cite{li2024learning} & 98.73 & 6.32 & 95.55 & 22.89 & 93.34 & 27.60 & 91.60 & 35.21 & 94.81 & 23.01 \\
 \midrule
  \multicolumn{11}{c}{\textbf{Zero-Shot Training-free Methods}} \\
Mahalanobis \cite{lee2018simple}      & 55.89 & 99.33 & 59.94 & 99.41 & 65.96 & 98.54 & 64.23 & 98.46 & 61.50 & 98.94 \\
Energy \cite{liu2020energy}           & 85.09 & 81.08 & 84.24 & 79.02 & 83.38 & 75.08 & 65.56 & 93.65 & 79.57 & 82.21 \\
MCM \cite{ming2022delving} & 94.59 & 32.20 & 92.25 & 38.80 & 90.31 & 46.20 & 86.12 & 58.50 & 90.82 & 43.93 \\
NegLabel \cite{jiang2024negative} &99.49&1.91&95.49& 20.53 & 91.64 & 35.59 & 90.22 & 43.56 & 94.21 & 25.40 \\
% nips_abla_ada_comb_fourood/imagenet_fixedclip_negoodprompt_test_ood_ood_clip_tta_ttaprompt_combine_beta5.5_thres0.5_gap0.5_memleng10_lambda1.0_group_num_5_random_True_ood_taskada
\rowcolor{HighLight} \textbf{AdaNeg (Ours)} & \textbf{99.71} & \textbf{0.59} & \textbf{97.44} & \textbf{9.50} & \textbf{94.55} & \textbf{34.34} & \textbf{94.93} & \textbf{31.27} & \textbf{96.66} & \textbf{18.92} \\
\bottomrule
\end{tabular}
\vspace{-0.2cm}
\end{table}

\begin{table}[tb] \small
    \centering
\caption{OOD detection results on the OpenOOD benchmark, where ImageNet-1k is adopted as ID dataset. Full results are available in Tab. \ref{tab:openood_imagenet_full_adaneg}.}\label{tab:openood_imagenet_adaneg}
% \vspace{-0.2cm}
\begin{tabular}{l|cc|cc|c}
\toprule
\multirow{2}{*}{ Methods } & \multicolumn{2}{|c|}{ FPR95 $\downarrow$} & \multicolumn{2}{|c|}{$\mathrm{AUROC} \uparrow$} & ACC $\uparrow$ \\
\cline{2-5}
& Near-OOD & Far-OOD & Near-OOD & Far-OOD & ID \\
\midrule
 \multicolumn{6}{c}{\textbf{Methods requiring training (or fine-tuning)}} \\
GEN \cite{liu2023gen}  & -- & -- & 78.97 & 90.98 & 81.59 \\
AugMix \cite{hendrycks2019augmix} + ReAct \cite{sun2021react} & -- & -- & 79.94 & 93.70 & 77.63 \\
RMDS \cite{ren2021simple} & -- & -- & 80.09 & 92.60 & 81.14  \\
SCALE \cite{xu2023scaling} & -- & -- & 81.36 & 96.53 &  76.18 \\
AugMix \cite{hendrycks2019augmix} + ASH \cite{djurisic2022extremely}   & -- & -- & 82.16 & 96.05 & 77.63 \\
LAPT \cite{zhang2024lapt} & 58.94 & 24.86 & 82.63 & 94.26 & 67.86 \\
\midrule
\multicolumn{6}{c}{\textbf{Zero-shot Training-free Methods}} \\
MCM \cite{ming2022delving}        & 79.02 & 68.54 & 60.11 & 84.77 & 66.28 \\
%%%%%%%%%%%%%%%%%%%%%%% LAPT results, last 1k texts, with overlap text.
% NegLabel \cite{jiang2024negative} & 68.18 & 27.34 & 76.92 & 93.30 & 66.82\\
% \rowcolor{HighLight} \textbf{LAPT (eccv)} & \textbf{58.94} & \textbf{24.86} & \textbf{82.63} & \textbf{94.26} & \textbf{67.86} \\

%%%%%%%%%%%%%  official NegLabel reimplement, 10K text, 100 group, dedup.
NegLabel \cite{jiang2024negative} & 69.45 & 23.73 & 75.18 & 94.85 & 66.82\\
% nips_abla_thres_gap/imagenet_fixedclip_negoodprompt_test_ood_ood_clip_tta_ttaprompt_combine_beta10.5_thres0.5_gap0.5_group_num_5_random_True_ood
\rowcolor{HighLight} \textbf{AdaNeg (Ours)} & \textbf{67.51} & \textbf{17.31} & \textbf{76.70} & \textbf{96.43} & \textbf{67.13} \\
%%%%%%  not dataset individual, utilize samples of same group (near good group)
% \rowcolor{HighLight} \textbf{AdaNeg (Ours)} & \textbf{65.41}$\pm$0.02 & \textbf{17.05} & \textbf{77.90} & \textbf{96.17} & \textbf{67.36}$\pm$0.08 \\
\bottomrule
\end{tabular}
\vspace{-0.1cm}
\end{table}

\subsection{Analyses and Discussions} \label{subsec:adaneg_analyses}

\begin{table}[tb] \small
    \centering
    \caption{OOD detection results with different score functions, where results are reported with ImageNet ID dataset under the OpenOOD setup.} \label{Tab:ablation_score_fun}
    \begin{tabular}{ccc|cc}
        \toprule
         $S_{nl}$ & $S_{ta}$ & $S_{sa}$        & Near-OOD AUROC $\uparrow$        & Far-OOD AUROC $\uparrow$  \\
         \midrule
         \Checkmark  & && 75.18 & 94.85  \\
          % & \Checkmark &  & 73.99 & 96.35 \\
          % &  &  \Checkmark & 74.46 & 96.20 \\
          & \Checkmark &  & 75.76 & 96.20 \\
          &  &  \Checkmark & 76.03 & 96.35 \\
          \midrule
        \Checkmark  & \Checkmark &  & 76.49 & 96.45  \\
                \Checkmark  &  & \Checkmark &  76.70 & 96.63 \\
         \bottomrule
    \end{tabular}
\end{table}

\begin{figure}[tb]
	\centering
	\begin{subfigure}{0.32\linewidth}
		\includegraphics[width=\linewidth]{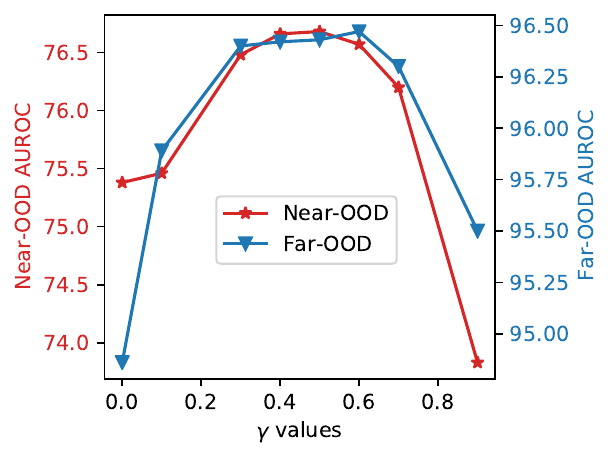}
		\caption{Threshold $\gamma$}
		\label{subfig:gamma_value}
	\end{subfigure}
	\hfill
	\begin{subfigure}{0.32\linewidth}
		\includegraphics[width=\linewidth]{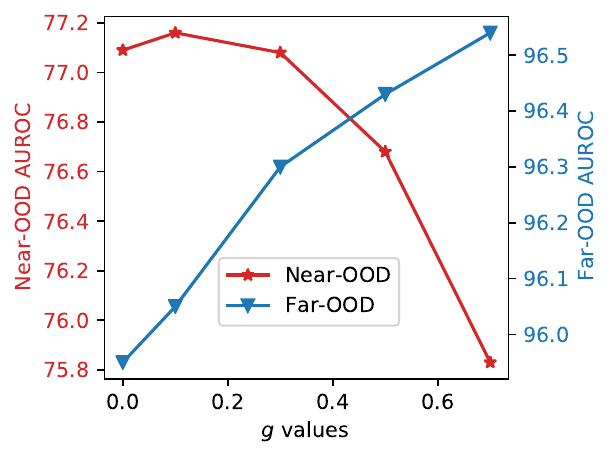}
		\caption{Gap value $g$}
		\label{subfig:gap_value}
	\end{subfigure}
	\hfill
	\begin{subfigure}{0.32\linewidth}
		\includegraphics[width=\linewidth]{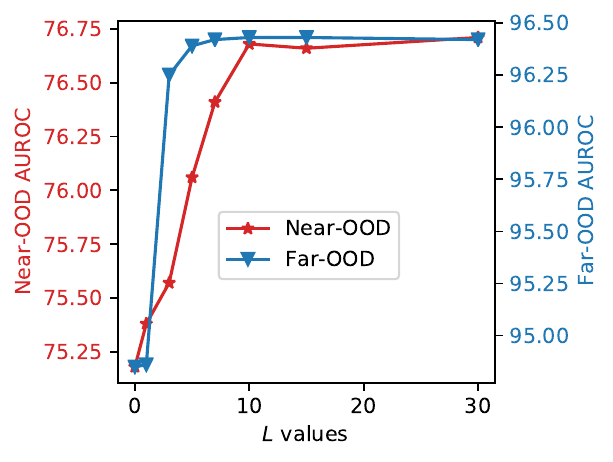}
		\caption{Memory length $L$}
		\label{subfig:memory_length}
	\end{subfigure}
%	\begin{subfigure}{0.3\linewidth}
%		\includegraphics[width=\linewidth]{images_adaneg/lambda_adaneg.pdf}
%		\caption{$\lambda$ value}
%		\label{subfig:lambda_value}
%	\end{subfigure}
%	\hfill
%	\begin{subfigure}{0.3\linewidth}
%		\includegraphics[width=\linewidth]{images_adaneg/beta_adaneg.pdf}
%	\caption{$\beta$ value}
%	\label{subfig:beta_value}
%	\end{subfigure}
	\caption{Analyses on the hyper-parameters of (a) threshold $\gamma$ in Eq. \ref{Equ:criterion_memorized_tau_g}, (b) gap value $g$ in Eq. \ref{Equ:criterion_memorized_tau_g}, and (c) memory length $L$ on the ImageNet dataset under OpenOOD setting.
	}
	\label{Fig:analyses_hyperparam_adaneg}
	\vspace{-0.1cm}
\end{figure}

\begin{table}[tbh]\footnotesize
	\centering
	\caption{Analyses on the time complexity of our AdaNeg and its competitors. `Training' measures the training time, and `Param.' presents the number of learnable parameters. `FPS' reflects the inference speed, measured with a batch size of 256. Results are achieved with a NVIDIA V100 GPU. 
		% 		% `GFLOPs' are calculated during training or test-time training with gradient back-propagation, sssss
	}
	\label{Tab:computation_efficiency_adaneg}
	\begin{tabular}{l|ccccc}
		\toprule
		\textbf{Methods} & \textbf{Training} & \textbf{FPS} &  \textbf{Param.} & \textbf{FPR95  $\downarrow$} \\
		\midrule
		% \textcolor{gray}{Zero-shot}  \\
		ZOC \cite{esmaeilpour2022zero}  & $>$24h  & 287 & 336M & 85.19 \\
		CLIPN \cite{wang2023clipn}    & $>$24h  & 605 & 37.8M & 31.10 \\
  		LoCoOp \cite{miyai2024locoop}   & 9h & 625       & 8K  & 28.66 \\
		\midrule
		MCM \cite{ming2022delving}     & -- & 625 & -- & 43.93 \\  %%% 1.6ms pure forward
		NegLabel \cite{jiang2024negative} & -- & 592 & -- & 25.40 \\  %% 1.8ms??
		% \rowcolor{HighLight} \textbf{LAPT (Ours)} & 1h & 10.5ms & 4K &  23.40 \\
		\rowcolor{HighLight} \textbf{AdaNeg (Ours)} & -- & 476 & -- &  18.92 \\  %%% 2.1ms
		% Others   & \\
		% CLIP \cite{radford2021learning}   & --  & 10.1ms & 0 & 0 & 60.33 \\
		% CALIP \cite{guo2023calip}   & -- & 10.2ms & 0 & 0 & 60.57 \\
		% TPT \cite{shu2022test}    & --  & 436ms            & $>$10& 0.01M & 60.74   \\
		% CoOp \cite{zhou2022learning}          & 14 h & 10.2ms & 0.01M  &62.95 \\
		\bottomrule
	\end{tabular}
\end{table}

\textbf{Score Functions.} 
As illustrated in Tab. \ref{Tab:ablation_score_fun}, the adaptive score functions $S_{ta}$ and $S_{sa}$ consistently outperforms the fixed $S_{nl}$, validating the advantage of adaptive proxies over fixed label proxies. The sample-adaptive score $S_{sa}$ slightly surpasses the task-adaptive one $S_{ta}$, verifying the usefulness of fine-grained sample characteristics. Combining adaptive image-based proxies and fixed text-based proxies leads to the best performance, proving their complementarity.

\textbf{Threshold $\gamma$ and gap $g$ in Eq. \ref{Equ:criterion_memorized_tau_g}.}
As illustrated in Fig. \ref{Fig:analyses_hyperparam_adaneg}, employing a moderate threshold $\gamma$ (\eg, 0.4 < $\gamma$ < 0.6) proves effective in distinguishing OOD samples across various scenarios. However, the efficacy of the gap parameter $g$ depends upon the specific characteristics of different OOD datasets. A larger $g$ facilitates the identification of more confidently classified ID/OOD samples, thereby improving the detection of far-OOD samples. Conversely, a smaller $g$ is advantageous for near-OOD detection as it accommodates low-confidence OOD samples, which are typically more prevalent in near-OOD scenarios.

In scenarios where the OOD distribution is entirely unknown, we adopt a conservative approach by setting $g$ to 0.5 in all experiments. This balanced setting provides a robust baseline for performance across a variety of conditions. However, if prior knowledge regarding the difficulty levels of OOD datasets is accessible, tailoring the hyperparameters—such as opting for a smaller $g$ in more challenging OOD contexts—can yield enhanced detection performance.

\textbf{ID-OOD imbalanced Test Data.}
To investigate the stability of our method with imbalanced ID-OOD test data, we construct test sets with various mixture ratios of ID and OOD samples. Specifically, we adopted the 1.28M ImageNet training data as ID and randomly sampled 12.8K and 1.28K instances from the SUN OOD dataset to construct the ID:OOD ratios of 100:1 and 1000:1 settings, respectively. To construct the ID:OOD ratios of 1:100, 1:10, 1:1, and 10:1 settings, we randomly sampled 40K samples from the SUN dataset as OOD and randomly sampled 400, 4K, 40K, and 400K instances from the ImageNet training data.  As shown in Tab. \ref{tab:adaneg_mixratio}, our method outperforms NegLabel across a wide range of mixture ratios (from 1:100 to 100:1), validating the robustness and reliability of our approach. However, unbalanced mixture ratios do pose a challenge to our method. Our approach performs the best in scenarios with a balanced mixture of ID and OOD samples, reducing the FPR95 by 11.18\%. As the mixture ratio becomes increasingly unbalanced, the improvement brought by our method gradually decreases. When the unbalanced ratio reaches 1000:1, our method shows some negative impact.

After a more detailed analysis, we discovered that the fundamental issue stems from the increased proportion of misclassified samples in the memory due to the growing ID-OOD imbalance. To effectively address this problem, we refine the selection criterion for memorizing OOD samples by adaptively adjusting the gap $g$. We refer to this adaptive gap strategy as AdaGap, which significantly improves the robustness of our method against ID-OOD imbalanced test data, as demonstrated in Table \ref{tab:adaneg_mixratio}.
To summarize, we first estimate the ratio of ID to OOD in the test data online. If there is a higher proportion of ID/OOD compared to OOD/ID, we adjust the standards for caching ID/OOD data into memory by dynamically modifying the gap in Eq. \ref{Equ:criterion_memorized_tau_g}. For more detailed implementation, please refer to the Appendix \ref{subsec:more_analyses_adaneg}.

% 经过更深入的分析，我们发现该问题的根源是随着ID-OOD imbalance 的增加，misclassified samples 在memory 中的占比增加导致的。 This issue can be effectively addressed by refining the selection criterion for OOD memorization by adaptively adjusting the gap g. We term this adaptive gap strategy design as AdaGap, which significantly enhances the robustness of our method against ID-OOD imbalanced test data, as shown in Tab. \ref{tab:adaneg_mixratio}.  简要来说，我们首先online 估计测试数据中ID 和 OOD 的比例。如果ID/OOD 比较多，就相应提高cache ID/OOD 数据到memory 的标准 by online 调整gap in Eq. (8). Please find more detailed implementation in the Appendix \ref{subsec:more_analyses_adaneg}.

\begin{table}[tb]
    \centering
     \caption{FPR95 ($\downarrow$) with different mixture ratios of ID and OOD samples.}
    \label{tab:adaneg_mixratio}
    \begin{tabular}{l|ccccccc}
    \toprule
    ID:OOD Ratio   & 1:100 & 1:10 & 1:1     & 10:1 & 100:1 & 1K:1  \\
    \midrule
    NegLabel        & 22.42 & 21.11  & 20.99 & 20.92 & 21.48 & 23.69 \\
    AdaNeg          & 21.00 & 12.49  & 9.81  & 15.61 & 20.71 & 26.28 \\
    AdaNeg (With AdaGap) & 20.50 & 12.22  & 9.73  & 12.98 & 15.61 & 18.43 \\
    \bottomrule
    \end{tabular}
\end{table}

\begin{table}[tb]
\centering
\caption{OOD detection results with BIMCV-COVID19+ \cite{vaya2020bimcv}, where a VITB/16 CLIP encoder is adopted.} \label{tab:adaneg_medical_exp}
\begin{tabular}{lcccc|cc}
\toprule
& \multicolumn{6}{c}{OOD datasets}  \\
\multicolumn{1}{c}{\multirow{2}{*}{Methods}} & \multicolumn{2}{c}{CT-SCAN} & \multicolumn{2}{c}{X-Ray-Bone} & \multicolumn{2}{c}{Average} \\ \cline{2-3} \cline{4-5} \cline{6-7} 
 & \small AUROC$\uparrow$ & \small FPR95$\downarrow$& \small AUROC$\uparrow$ & \small FPR95$\downarrow$ & \small AUROC$\uparrow$ & \small FPR95$\downarrow$        \\
 \midrule
NegLabel  & 63.53 & 100 & 99.68 & 0.56 & 81.61 & 50.28 \\
\textbf{AdaNeg (Ours)} & \textbf{93.48} & 100 & \textbf{99.99} & \textbf{0.11} & \textbf{96.74} & \textbf{50.06} \\
\bottomrule
\end{tabular}
\end{table}

\textbf{Generalization to Other Domains.}
Besides experiments with natural images, we further validate our method on the BIMCV-COVID19+ dataset \cite{vaya2020bimcv}, which includes medical images, following the OpenOOD setup. Specifically, we select BIMCV as the ID dataset, which includes chest X-ray images CXR (CR, DX) of COVID-19 patients and healthy individuals. For the OOD datasets, we follow the OpenOOD setup and use CT-SCAN and X-Ray-Bone datasets. The CT-SCAN dataset includes computed tomography (CT) images of COVID-19 patients and healthy individuals, while the X-Ray-Bone dataset contains X-ray images of hands.
As illustrated in Tab. \ref{tab:adaneg_medical_exp}, our AdaNeg method consistently outperforms NegLabel on this medical image dataset.

\textbf{Memory Length.}
As demonstrated in Fig. \ref{subfig:memory_length}, there is a positive correlation between the memory length $L$ and the performance outcomes, affirming the efficacy of feature memorization from another perspective. 
In all our experiments, we have set $L$ to 10.
% \begin{table}[h!]
%     \centering
%     \begin{tabular}{c|ccccccccc}
%     \hline
%         Memory Length $L$ &  0 & 1 & 3 & 5 & 10 & 15 &30 & 50  \\
%         near AUROC & 75.18 & 75.39 & 75.76 & 76.05 & 76.32 & 76.27 & 76.26\\
%         far AUROC & 94.85 & 94.86  & 96.24 & 96.26 & 96.26 & 96.23 & 96.22 \\
%         \hline
%     \end{tabular}
%     \caption{Memory length, turn it to figure. }
% \end{table} 

%\textbf{Positive vs. Negative Feature Caching.}
%%% 现在是把positive 和 negative 的图像都cache了; 但是应该negative image 对OOD detection 的贡献更大一些； positive image 应该对ID classification 贡献多一些。 需要验证。

\textbf{Complexity Analyses.}
As analyzed in Table \ref{Tab:computation_efficiency_adaneg}, our AdaNeg approach does not introduce any learnable parameters or require model training. Furthermore, it significantly enhances performance while maintaining a fast testing speed.

More analyses and discussions on the $\lambda$ in Eq. \ref{equ:adaneg_score_all}, $\beta$ in Eq. \ref{Equ:sample_adaptive_classifier}, various backbone architectures, the ordering of test samples, complementarity to training-based method, and the number of test data can be found in Appendix \ref{subsec:more_analyses_adaneg}.

% More analyses and discussions on the $\lambda$ in Eq. \ref{equ:adaneg_score_all}, $\beta$ in Eq. \ref{Equ:sample_adaptive_classifier}, various backbone architectures, the ordering of test samples, and number of test data can be found in Appendix \ref{subsec:more_analyses_adaneg}.

\section{Conclusion and Limitations} \label{sec:conclusion_adaneg}

We proposed adaptive negative proxies that aligned more effectively with OOD distributions, thereby providing more effective guidance for OOD detection.
These proxies were constructed using a task-aware feature memory that selectively cached discriminative image features during testing. Our approach facilitated the generation of both task-adaptive and sample-adaptive proxies through carefully designed memory reading operations. Notably, our method was training-free and annotation-free, and it maintained fast testing speed and achieved state-of-the-art results on various benchmarks.
These results validated the effectiveness of the proposed adaptive proxies.

One minor limitation of our method is the introduction of an external memory requirement. For example, this memory occupies a storage space of 214.75MB when using the ImageNet dataset as the ID, which may pose challenges for storage-constrained applications.

{\small
\bibliographystyle{plain}
\bibliography{egbib}
}

%%%%%%%%%%%%%s%%%%%%%%%%%%%%%%%%%%%%%%%%%%%%%%%%%%%%%%%%%%%%%
\newpage
\appendix

\renewcommand{\thetable}{A\arabic{table}}
\renewcommand{\thefigure}{A\arabic{figure}}
\numberwithin{equation}{section}

\section{Appendix}

\subsection{ID-Similarity to OOD Ratio with ground truth ID and OOD labels} \label{subsec:id_aligning_scores}
We introduce the ID-Similarity to OOD Ratio (ISOR) to quantitatively measure the relative alignment of negative proxies with ground truth OOD labels, compared to their alignment with ID labels. In implementation, we adapt the score function of NegLabel (\ie, Eq. \ref{equ:neglabel_score}), which originally measures the similarity of a test image to ID labels over negative labels. We modify this function by replacing the negative labels with ground truth OOD labels and changing the input from test images to negative proxies (\eg, negative texts), while keeping other aspects consistent with Eq. \ref{equ:neglabel_score}. This modified score function allows us to assess the degree of similarity between the inputs and ground truth ID/OOD labels, thereby quantifying the relative alignment between negative proxies and OOD labels. Specifically, lower ISOR indicates a higher similarity to OOD labels and a reduced similarity to ID labels.

\subsection{Entropy-based caching strategy for full memory} \label{subsec:entropy_based_caching}
The memory we construct is of finite length; hence, as the number of cached images increases, it may become fully occupied. To address this, we have devised a simple strategy to retain only those images with low entropy, \eg, high confidence. Specifically, when storing an image in memory, we also record its entropy pertinent to OOD detection:
\begin{align}
 \mathrm{Entropy}(\vv) = - S_{nl}(\vv) \log(S_{nl}(\vv))  - (1-S_{nl}(\vv) )\log(1 -S_{nl}(\vv)),
\end{align}
where $S_{nl}(\vv)$ represents the probability of belonging to the ID, as shown in Eq. (\ref{equ:neglabel_score}).
Given the entropy of the current test image and a full memory $\mathbf{M}_y$, we replace the item with the maximum entropy in $\mathbf{M}_y$ with the current image feature if the current test image exhibits lower entropy. Otherwise, we do not cache the current test sample.

\subsection{Evaluation criteria} \label{subsec:criterion_definition}
Following common practice \cite{huang2021mos,ming2022delving,jiang2024negative}, we employ the following three evaluation metrics: (1) FPR95, which measures the false positive rate for OOD samples when the detection accuracy for ID samples is at 95\%; (2) AUROC, the area under the receiver operating characteristic curve; and (3) ID ACC, which quantifies the accuracy of correctly identifying and classifying ID samples.

\subsection{Detailed results on ImageNet dataset under OpenOOD setting}
The detailed OOD detection results on the OpenOOD benchmark are presented in Tab. \ref{tab:openood_imagenet_full_adaneg}.

\subsection{Results on CIFAR10/100} \label{subsec:cifar_results}
As illustrated in Tab. \ref{tab:openood_cifar_adaneg}, the advantage of our AdaNeg also holds on the CIFAR10/100 dataset. Notably, our method achieves new state-of-the-art results in the far-OOD setting under a zero-shot training-free manner, even outperforming its competitors training on the full labeled training set.

\begin{table}[tb]
    \centering
\caption{Detailed OOD detection results on the OpenOOD benchmark, where ImageNet-1k is adopted as ID dataset.}\label{tab:openood_imagenet_full_adaneg}
\begin{tabular}{ll|c|c}
\toprule
Near-/Far-OOD & Datasets & FPR95 $\downarrow$ & AUROC $\uparrow$\\
\midrule
\multirow{3}{*}{Near-OOD} & SSB-hard \cite{vaze2021open} &  74.91 & 75.11  \\
                          & NINCO \cite{bitterwolf2023ninco}   &  60.10 & 78.30   \\
                          & \textbf{Mean}                         &  67.51 & 76.70  \\
 \midrule     
\multirow{4}{*}{Far-OOD} & iNaturalist \cite{van2018inaturalist} & 0.72 &  99.72 \\
                          & Textures   \cite{cimpoi2014describing} & 21.40 &  95.71 \\
                          & OpenImage-O  \cite{wang2022vim} & 29.81 & 93.87 \\
                          & \textbf{Mean}                            & 17.31 & 96.43 \\
\bottomrule
\end{tabular}
\vspace{-0.2cm}
\end{table}

% \begin{table}[tb]
%     \centering
% \caption{Detailed full-spectrum OOD detection results of our method on the OpenOOD benchmark, where ImageNet-1k is adopted as ID dataset.}\label{tab:openood_lapt_fsood_full}
% \vspace{-0.2cm}
% \begin{tabular}{ll|c|c}
% \toprule
% % \multirow{2}{*}{ Methods } & \multicolumn{2}{|c|}{ FPR95 $\downarrow$} & \multicolumn{2}{|c|}{$\mathrm{AUROC} \uparrow$} & ACC $\uparrow$ \\
% % \cline{2-5}
% Near-/Far-OOD & Datasets & FPR95 $\downarrow$ & AUROC $\uparrow$\\
% \midrule
% \multirow{3}{*}{Near-OOD} & SSB-hard \cite{vaze2021open} &  xxx & xxx  \\
%                           & NINCO \cite{bitterwolf2023ninco} & xxx & xxx    \\
%                           & \textbf{Mean}     &   xx & xx \\
%  \midrule     
% \multirow{4}{*}{Far-OOD} & iNaturalist \cite{van2018inaturalist} & xxx & xxx \\
%                           & Textures   \cite{cimpoi2014describing} & xxx & xxx \\
%                           & OpenImage-O  \cite{wang2022vim} & xxx & xxx \\
%                           & \textbf{Mean}     &  xx & xx \\
% % \rowcolor{HighLight} \textbf{LAPT (Ours)} & \textbf{58.94} & \textbf{24.86} & \textbf{82.63} & \textbf{94.26} & \textbf{67.86} \\
% \bottomrule
% \end{tabular}
% \vspace{-0.2cm}
% \end{table}
% % Optionally include supplemental material (complete proofs, additional experiments and plots) in appendix.
% % All such materials \textbf{SHOULD be included in the main submission.}

\begin{table}[tb] \small
	\centering
	\caption{OOD detection results with CIFAR10/100 on the OpenOOD benchmark. Full results are provided in Tables \ref{tab:openood_cifar10_full_adaneg} and \ref{tab:openood_cifar100_full_adaneg}.} 
	\label{tab:openood_cifar_adaneg}
	\centering
	\begin{subtable}[t]{0.98\textwidth}
		\centering
		\begin{tabular}{l|cc|cc}
			\toprule
			\multirow{2}{*}{ Methods } & \multicolumn{2}{|c|}{ FPR95 $\downarrow$} & \multicolumn{2}{|c}{$\mathrm{AUROC} \uparrow$}  \\
			\cline{2-5}
			& Near-OOD & Far-OOD & Near-OOD & Far-OOD \\
			\midrule
			\multicolumn{5}{c}{\textbf{Methods requiring training (or fine-tuning)}} \\
			PixMix \cite{hendrycks2022pixmix} + KNN \cite{sun2022out} & -- & -- &  93.10 & 95.94  \\
			OE \cite{hendrycks2018deep} + MSP \cite{hendrycks2016baseline} &-- & -- &  94.82 & 96.00  \\
			PixMix \cite{hendrycks2022pixmix} + RotPred \cite{hendrycks2019using}  & -- & -- &94.86 &98.18  \\
			\midrule
			\multicolumn{5}{c}{\textbf{Zero-shot Training-free Methods}} \\
			MCM \cite{ming2022delving} &30.86 & 17.99 & 91.92 &  95.54 \\
			NegLabel \cite{jiang2024negative} &  28.75 & 6.60 & 94.58 & 98.39  \\
			\rowcolor{HighLight} \textbf{AdaNeg (Ours)} & \textbf{20.40} & \textbf{2.79} & \textbf{94.78} & \textbf{99.26} \\
			\bottomrule
		\end{tabular}
		\caption{CIFAR10 as ID dataset}
	\end{subtable}
	\begin{subtable}[t]{0.98\textwidth}
		\centering
		\begin{tabular}{l|cc|cc}
			\toprule
			\multirow{2}{*}{ Methods } & \multicolumn{2}{|c|}{ FPR95 $\downarrow$} & \multicolumn{2}{|c}{$\mathrm{AUROC} \uparrow$}  \\
			\cline{2-5}
			& Near-OOD & Far-OOD & Near-OOD & Far-OOD  \\
			\midrule
			\multicolumn{5}{c}{\textbf{Methods requiring training (or fine-tuning)}} \\
			GEN \cite{liu2023gen}  & -- & -- & 81.31 & 79.68  \\
			VOS \cite{du2022vos} + EBO \cite{liu2020energy} & -- & -- & 80.93 & 81.32  \\
			SCALE \cite{xu2023scaling} &-- & -- & 80.99 & 81.42  \\
			OE \cite{hendrycks2018deep} + MSP \cite{hendrycks2016baseline} & -- & -- & 88.30 & 81.41  \\
			\midrule
			\multicolumn{5}{c}{\textbf{Zero-shot Training-free Methods}} \\
			MCM \cite{ming2022delving}  & 75.20 & 59.32 & 71.00 & 76.00 \\
			NegLabel \cite{jiang2024negative} & 71.44 & 40.92  & 70.58 &  89.68  \\
			\rowcolor{HighLight} \textbf{AdaNeg (Ours)} & \textbf{59.07} & \textbf{29.35} & \textbf{84.62} & \textbf{95.25}   \\
			\bottomrule
		\end{tabular}
		\caption{CIFAR100 as ID dataset}
	\end{subtable}
	\vspace{-0.2cm}
\end{table}

\begin{table}[tb]
    \centering
\caption{Detailed OOD detection results of on the OpenOOD benchmark, where CIFAR100 is adopted as ID dataset.}\label{tab:openood_cifar100_full_adaneg}
\begin{tabular}{ll|c|c}
\toprule
Near-/Far-OOD & Datasets & FPR95 $\downarrow$ & AUROC $\uparrow$\\
\midrule
\multirow{3}{*}{Near-OOD} & CIFAR10 \cite{krizhevsky2009learning} &  58.24 & 79.91  \\
                          & TIN \cite{le2015tiny}   &  59.90 & 89.34   \\
                          & \textbf{Mean}  & 59.07 & 84.62  \\
 \midrule     
\multirow{5}{*}{Far-OOD} & MNIST \cite{deng2012mnist} & 4.18 &  97.90 \\
                          & SVHN \cite{netzer2011reading}    & 6.03 &  97.60 \\
                          & Texture \cite{cimpoi2014describing}   & 30.00 & 95.14  \\
                          & Places365 \cite{zhou2017places} & 77.20 & 90.35  \\
                          & \textbf{Mean}  & 29.35 & 95.25 \\
\bottomrule
\end{tabular}
\vspace{-0.2cm}
\end{table}

\begin{table}[tb]
    \centering
\caption{Detailed OOD detection results of our method on the OpenOOD benchmark, where CIFAR10 is adopted as ID dataset.}\label{tab:openood_cifar10_full_adaneg}
\begin{tabular}{ll|c|c}
\toprule
Near-/Far-OOD & Datasets & FPR95 $\downarrow$ & AUROC $\uparrow$\\
\midrule
\multirow{3}{*}{Near-OOD} & CIFAR100 \cite{krizhevsky2009learning} &  35.80 & 90.93  \\
                          & TIN \cite{le2015tiny}   &  5.01 & 98.63   \\
                          & \textbf{Mean}  & 20.40 & 94.78  \\
 \midrule     
\multirow{5}{*}{Far-OOD} & MNIST \cite{deng2012mnist}   & 0.13 &  99.96 \\
                          & SVHN \cite{netzer2011reading}    & 0.04 &  99.97 \\
                          & Texture \cite{cimpoi2014describing}  & 0.04 & 99.82 \\
                          & Places365 \cite{zhou2017places} & 10.93 & 97.29  \\
                          & \textbf{Mean} & 2.79 & 99.26  \\
\bottomrule
\end{tabular}
\vspace{-0.2cm}
\end{table}

\subsection{More analyses} \label{subsec:more_analyses_adaneg}

\textbf{Detailed Implementation of AdaGap Module.}
We have implemented an adaptive gap (AdaGap) module to adjust the memorization selection criteria dynamically. This strategy builds on the observation that as the score $S_{nl}$ increases/decreases, the probability that a sample is ID/OOD also increases accordingly. By enforcing a stringent selection criterion, we can effectively minimize the inclusion of misclassified samples in our memory. Specifically, we first online estimate the ratio of ID to OOD samples in the test data using a First-In-First-Out queue, which caches the ID/OOD estimation (cf. Eq. \ref{Equ:criterion_memorized_tau_g}) of the most recent $N$ samples:
\begin{equation}
    MR = \frac{\text{Estimated ID Number}}{\text{Estimated ID Number + Estimated OOD number}},
\end{equation}
where the ID and OOD numbers are acquired within the queue. 

Leveraging the estimated mix ratio ($MR$), we can dynamically adjust the gap 
$g$ in memory caching to avoid a majority of misclassified samples within the memory. For instance, if we find that ID samples constitute the majority of the test samples (\ie, $MR$ > 0.5), this could lead to an increased proportion of ID samples in the OOD memory. In response, we can adjust the selection criterion for OOD memorization to only cache higher confidence OOD samples. This adjustment is achieved by modifying the selection criterion for memorization in Eq. \ref{Equ:criterion_memorized_tau_g} as follows:

\begin{align}
    \mathrm{Negative}:&  S_{nl}(\vv) < \gamma - g\gamma \to S_{nl}(\vv) < \gamma - \max(g, MR)\gamma  , \notag  \\
    \mathrm{Positive}:&  S_{nl}(\vv) \geq  \gamma + g(1-\gamma) \to  S_{nl}(\vv) \geq  \gamma +  \max(g,1-MR)(1-\gamma) , 
\end{align}  

where $g =0.5$ is the default gap analyzed in Figure \ref{subfig:gap_value}. 
In this way, under ID/OOD balanced conditions (\ie, $MR$ = 0.5), our method aligns with our original version. However, if the proportion of ID samples is higher in the test samples' estimation (\ie, $MR$ > 0.5), we raise the standard for storing negative samples in the memory. In an extreme case when $MR$ = 1, we estimate that there might be no OOD samples among the test samples; thus, we stop storing test samples in the negative memory and only selectively cache test samples into the positive memory. We adjust our approach conversely when the $MR$ value is lower than 0.5.

Please note that the selection criterion is dynamically adjusted online because the $MR$ is estimated with the most recent $N$ test samples. Here, we set  $N=10,000$ by default. This dynamic adjustment ensures that our memory caching strategy remains responsive to the evolving nature of the test sample distribution, thereby optimizing memory utilization and enhancing the accuracy of our domain distinction process.

\begin{figure}[tb]
	\centering
		\begin{subfigure}{0.4\linewidth}
			\includegraphics[width=\linewidth]{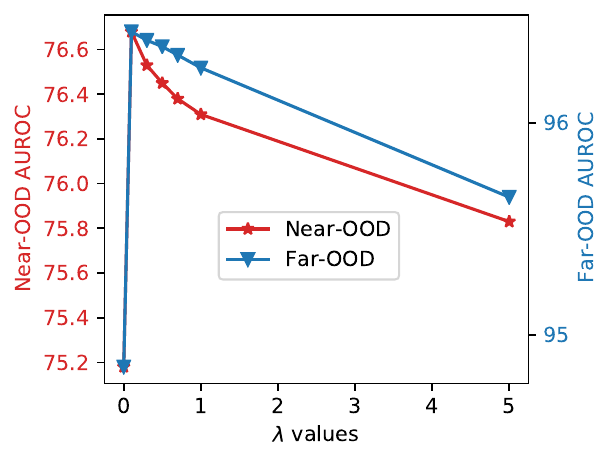}
			\caption{$\lambda$ value}
			\label{subfig:lambda_value}
		\end{subfigure}
		\begin{subfigure}{0.4\linewidth}
			\includegraphics[width=\linewidth]{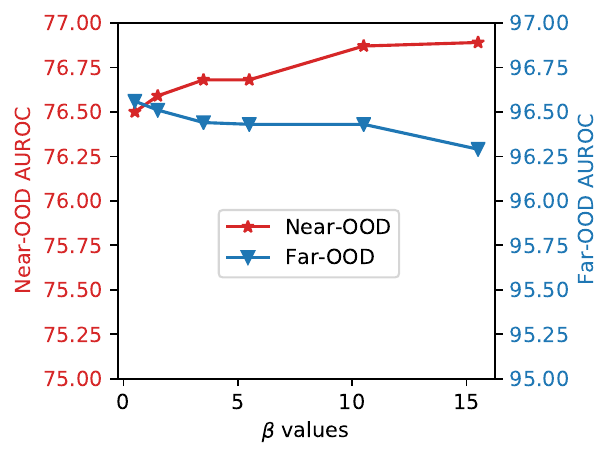}
		\caption{$\beta$ value}
		\label{subfig:beta_value}
		\end{subfigure}
	\caption{Analyses on the hyper-parameters of (a) $\lambda$ in Eq. \ref{equ:adaneg_score_all} and  (b) $\beta$ in Eq. \ref{Equ:sample_adaptive_classifier} on the ImageNet dataset under OpenOOD setting.
	}
	\label{Fig:analyses_hyperparam_adaneg_part2}
	\vspace{-0.2cm}
\end{figure}

\textbf{$\lambda$ in Eq. \ref{equ:adaneg_score_all}.}
As illustrated in Fig. \ref{subfig:lambda_value}, the OOD detection performance remains robust across a wide range of $\lambda$ values, \eg, from 0.01 to 1. For all experiments, we have set $\lambda$ to 0.1.

\textbf{$\beta$ in Eq. \ref{Equ:sample_adaptive_classifier}.}
Results with different $\beta$ values are shown in Fig. \ref{subfig:beta_value}, where OOD detection performance is robust to the $\beta$ values. We adopt $\beta$=5.5 in all experiments. 

%\textbf{$\beta$ in Eq. \ref{Equ:sample_adaptive_classifier}}
%
%%% 控制 加权系数的sharpness; 设定为5.5， 好像影响不大
%\begin{table}[h!]
%    \centering
%    \begin{tabular}{c|ccccccccc}
%    \hline
%         $\beta$ &   0.1 & 0.5 & 1.5 & 5.5 & 10.5 & 15.5 & 20.5 \\
%        near AUROC & 76.19 & 76.19 & 76.21 & 76.23 & 76.25 & 76.38  \\
%        far AUROC & 96.40 & 96.40 & 96.33  & 96.25 & 96.21 & 96.19 \\
%        \hline
%    \end{tabular}
%    \caption{ turn it to figure. }
%\end{table} 

% \textbf{group number}
%%% 不同任务最优的不同。 cifar 设成1 最好， 而ImageNet 好像5， 10 这样比较好。 这里不展开解释，略过。s

%\textbf{$\lambda$ in Eq. \ref{equ:adaneg_score_all}.}
%As illustrated in Fig. XX, the OOD detection performance remains robust across a wide range of $\lambda$ values, e.g., from 0.01 to 1. For all experiments, we have set $\lambda$ to 0.01.
% \begin{table}[h!]
%     \centering
%     \begin{tabular}{c|ccccccccc}
%     \hline
%         $\lambda$ & 0&  0.001 & 0.01 & 0.1 & 0.5 & 1 & 5 & 10 & 50 \\
%         near AUROC & 75.18& 75.76 & 76.95 & 76.72 & 76.41 & 76.26 & 76.13 & 76.06 & 75.99 \\
%         far auroc  &94.85 & 96.23 & 96.40 & 96.38 & 96.31 & 96.22 & 96.19 & 96.13 & 96.11 \\
%         \hline
%     \end{tabular}
%     \caption{Memory length, turn it to figure. }
% \end{table} 

\textbf{Ordering of Testing Data.}
Our AdaNeg selectively caches test data into memory, potentially causing variations in the results depending on the ordering of the test data. To rigorously test this aspect, we randomly shuffled the order of the test data using three distinct seeds. We observed that our method exhibits robustness to changes in the ordering of test data. Specifically, across three experiments conducted on the ImageNet dataset, the AUROC scores were 96.65\%, 96.69\%, and 96.64\%, respectively, demonstrating fluctuations of less than 0.1\%. We report the average results from three random runs in our paper.

\textbf{Various Backbone Architectures.}
Results with various VLMs architectures are illustrated in Tab. \ref{tab:ood_various_arch_adaneg}, where better results are typically achieved with stronger backbones.

\begin{table}[h!] \scriptsize
	\centering
	\caption{OOD detection results of our AdaNeg with different VLMs architectures, where ImageNet-1K is used as the ID dataset.} \label{tab:ood_various_arch_adaneg}
	\begin{tabular}{lcccccccc|cc}
		\toprule
		\multicolumn{11}{c}{OOD datasets}  \\
		\multicolumn{1}{c}{\multirow{2}{*}{Backbone}} & \multicolumn{2}{c}{INaturalist} & \multicolumn{2}{c}{Sun} & \multicolumn{2}{c}{Places} & \multicolumn{2}{c}{Textures} & \multicolumn{2}{c}{Average} \\ \cline{2-3} \cline{4-5} \cline{6-7} \cline{8-9} \cline{10-11}
		& \tiny AUROC$\uparrow$ & \tiny FPR95$\downarrow$& \tiny AUROC$\uparrow$ & \tiny FPR95$\downarrow$& \tiny AUROC$\uparrow$ & \tiny FPR95$\downarrow$& \tiny AUROC$\uparrow$ & \tiny FPR95$\downarrow$ & \tiny AUROC$\uparrow$ & \tiny FPR95$\downarrow$        \\
		\midrule
		%    \multicolumn{11}{c}{\textbf{ID dataset: ImageNet-1K}} \\
		% NegLabel \cite{jiang2024negative} &99.49&1.91&95.49& 20.53 & 91.64 & 35.59 & 90.22 & 43.56 & 94.21 & 25.40 \\ss
		ResNet50 & 99.58 & 1.18 & 97.37 & 10.56 & 93.84 & 43.19 & 94.18 & 35.00 & 96.24 & 22.48 \\
		% ResNet101& 99.62 & 1.00 & 97.62 & 8.93 & 93.69 & 42.90 & 92.88 &46.41 & 95.95 & 24.81 \\
		VITL/32 & 99.59 & 1.02 & 97.53 & 9.63 & 93.99 & 38.45 & 94.21 & 37.92 & 96.33 & 21.76 \\
		VITB/16 & 99.71 & 0.59 & 97.44 & 9.50 & 94.55 & 34.34 & 94.93 & 31.27 & 96.66 & 18.92 \\
		\bottomrule
	\end{tabular}
\end{table}

\textbf{Complementarity to Training-based Method.}
We validated the complementarity between our AdaNeg method and the latest works, NegPrompt \cite{li2024learning} and LAPT \cite{zhang2024lapt}, which use learned prompts. As shown in Table \ref{tab:adaneg_complementary_learning}, our method significantly improves performance over these approaches, demonstrating its complementarity with training-based methods.

\begin{table}[h!]
    \centering
        \caption{FPR95 ($\downarrow$) with the ID dataset of ImageNet. }
    \label{tab:adaneg_complementary_learning}
    \begin{tabular}{l|cccc|c}
    \toprule
    Methods   & INaturalist & SUN & Places & Textures & Average \\
    \midrule
    NegPrompt \cite{li2024learning} & 6.76 &23.41 & 28.32 & 34.57 & 23.27 \\
    + \textbf{AdaNeg} & 3.87 & 11.35 & 25.45 & 29.79 & 17.62 \\
    \midrule
    LAPT \cite{zhang2024lapt}      & 1.10 & 20.59 & 35.38 & 40.11 & 24.29  \\
    + \textbf{AdaNeg}  & \textbf{0.58} & \textbf{9.98}  & \textbf{30.47} & \textbf{24.25} & \textbf{16.32}  \\
    \hline
    \end{tabular}
\end{table}

\textbf{Number of Testing Data.}
We examine the dependency of our approach on the number of test images by evaluating its performance across different scales of test samples. 
As the number of test samples increases (from 900 to 90K), the cached feature data also increases, leading to an improvement in our method's results, as shown in Tab. \ref{tab:adaneg_test_num}. Even with a small number of test samples (\eg, 90 and 900), our method significantly reduces FPR95 compared to NegLabel, demonstrating its robustness across different numbers of test images.

Note that with only 90 test images, the task of distinguishing between ID and OOD samples degenerates into a simpler task since the number of test images is even smaller than the number of classes (\eg, 1000 for ImageNet). Consequently, both NegLabel and our method achieve lower FPR95 in such an easier scenario.

\begin{table}[h!]
    \centering
    \caption{FPR95 ($\downarrow$) with different numbers of test images, where test samples are randomly sampled from ImageNet (ID) and SUN (OOD) datasets, and we maintain a consistent ratio of ID to OOD samples at 5:4 throughout the experiments. }
    \label{tab:adaneg_test_num}
    \begin{tabular}{c|ccccc}
    \hline
    Num. of Test Images   & 90 & 900 & 9K & 45K &  90K \\
    \hline
    NegLabel & 14.00 & 20.44 & 20.71 & 20.51 & 20.53 \\
    AdaNeg   & \textbf{6.00}  & \textbf{10.12} & \textbf{9.78}  & \textbf{9.66}  &  \textbf{9.50} \\
    \hline
    \end{tabular}
\end{table}

\end{document}